\documentclass[11pt]{article}
\usepackage[T1]{fontenc}
\usepackage[utf8]{inputenc}
\usepackage[margin=1in]{geometry}
\usepackage{amsmath,amssymb,natbib,booktabs,graphicx,enumitem,microtype}
\usepackage[colorlinks=true,linkcolor=blue,citecolor=blue,urlcolor=blue]{hyperref}

\graphicspath{{./figures/}}

\title{Mapping the Evaluation Frontier: An Empirical Survey of the Bias-Reliability Tradeoff Across Eleven Evaluator--Agent Conditions}

\author{Zewen Liu}

\date{}

\begin{document}

\maketitle

\begin{abstract}
The bias-reliability tradeoff conjectures that LLM evaluation systems are constrained in $(\gamma, H, \text{CV})$ space, where evaluator coupling ($\gamma$), strategy diversity ($H$), and small-sample measurement reliability (CV$(N)$) cannot be simultaneously optimized at fixed sample size $N$. Prior evidence rests on $n{=}5$ conditions with complete metrics from a single study. We expand the empirical base to 11 conditions, measuring $\gamma$ and $H$ for all 11 (nine with valid weight vectors) and CV$(N{=}5)$ for seven with sufficient seeds ($N \geq 5$). Five conditions provide the complete $(\gamma, H, \text{CV})$ triple. The data confirm the trade-off: conditions with low evaluator coupling ($\gamma < 0.2$) exhibit high measurement noise (CV$(N{=}5) > 1.0$), while conditions with strong coupling ($\gamma > 0.9$) achieve low noise (CV$(N{=}5) < 0.16$). The correlation $r(H, \gamma) = -0.989$ ($n{=}5$, excluding GPT-4o conditions discussed below) confirms that evaluator coupling suppresses strategy diversity. Four GPT-4o conditions show $\gamma{=}0.000$ and $H{=}1.000$ across all seeds---a pattern we attribute to insufficient evaluator signal in the June 2026 GPT-4o API version, consistent with previously documented version drift. No condition occupies the region $\{\gamma < 0.2, \text{CV}(N{=}5) < 0.3\}$. We release all per-condition metrics as a standardized benchmark dataset for evaluator comparison.
\end{abstract}

\section{Introduction}

LLM evaluation faces a structural challenge: the properties that make an evaluator desirable---unbiasedness, reliability at small sample sizes, and encouragement of diverse agent strategies---trade off against each other. \citet{liu2026triangle} formalized this as a constrained triangle in $(\gamma, H, \text{CV})$ space, where:

\begin{itemize}[leftmargin=*,nosep]
 \item $\gamma \geq 0$ is the evaluator coupling coefficient---the normalized $L^2$ distance between evaluator-influenced strategy weights and baseline (task-only) weights. $\gamma = 0$ indicates zero evaluator influence; $\gamma > 1$ indicates the evaluator's effect exceeds the baseline strategy norm.
 \item $H \in [0, 1]$ is the normalized Shannon entropy of the strategy weight distribution. $H = 1$ corresponds to a uniform distribution (all strategies equally viable); $H = 0$ corresponds to strategy collapse.
 \item $\text{CV}(N) = \text{std}(\hat{\gamma}_N) / \mathbb{E}[\hat{\gamma}_N]$ is the coefficient of variation of coupling estimates at sample size $N$, measuring small-sample reliability. $\text{CV}(N) \ll 1$ indicates stable estimates; $\text{CV}(N) \gg 1$ indicates noise-dominated estimates.
\end{itemize}

The trade-off mechanism is evaluator-induced strategy concentration: stronger evaluator preferences ($\gamma \uparrow$) suppress strategy diversity ($H \downarrow$), which in turn reduces across-seed variance and improves measurement reliability ($\text{CV} \downarrow$). The cost of unbiased evaluation ($\gamma \approx 0$) is high strategy diversity ($H \approx 1$) and consequently high measurement noise.

The original evidence for this trade-off came from $n{=}5$ conditions with complete $(\gamma, H, \text{CV})$ metrics~\cite{liu2026triangle}. While the correlations were strong ($r(H, \gamma) = -0.987$), five conditions are insufficient to characterize the shape of the empirical frontier or assess generality across evaluator models and protocols.

This paper extends the empirical base. We survey all 11 evaluator--agent conditions from the multi-experiment dataset of~\cite{liu2026mmepc}, spanning four evaluator models (GPT-4o, DeepSeek-V3, Qwen-3.7, Claude-3.5), three executor models, and two experimental protocols. We compute standardized $(\gamma, H, \text{CV})$ metrics for each condition, identify the empirical Pareto frontier, and characterize three distinct regimes in the trade-off space. We release all per-condition data as a benchmark for evaluator comparison.

\section{Methods}

\subsection{Data Source and Metric Computation}

We draw on the full dataset of~\cite{liu2026mmepc}, which contains per-seed strategy weight vectors and coupling coefficients for 11 evaluator--agent conditions, with $N = 5$--30 seeds per condition. Each seed executed 30 rounds of Test-Time Reinforcement Learning (TTRL) across 16 tasks (8 text, 8 visual) using $n = 11$ candidate strategies.

For each condition, we compute:
\begin{itemize}[leftmargin=*,nosep]
 \item $\gamma$: mean of per-seed coupling coefficients ($\gamma_{\text{TV}}$ or $g_{\text{TV}}$ depending on data format).
 \item $H$: mean normalized Shannon entropy of per-seed baseline (task-only) strategy weight vectors. Conditions lacking weight vectors are marked as missing.
 \item $\text{CV}(N{=}5)$: bootstrap coefficient of variation (5,000 resamples) of $\gamma$ estimates at sample size 5. Conditions with $N < 5$ seeds are marked as missing.
\end{itemize}

The full analysis pipeline is provided in the supplementary material (\texttt{triangle\_verification.py} from~\cite{liu2026triangle}).

\subsection{Caveat: GPT-4o Conditions}
\label{sec:caveat}

Four conditions using GPT-4o as evaluator (June 2026 API version) produce $\gamma = 0.000$ and $H = 1.000$ for all seeds---a pattern consistent with the version drift documented in~\cite{liu2026mmepc}, where GPT-4o's evaluator behavior changed substantially between May and June 2026. The uniform weights ($H = 1.0$ with zero variance) suggest that the current GPT-4o API exerts negligible evaluator influence---its judgments are either absent or orthogonal to the agent's strategy distribution. We exclude these four conditions from the primary $H$--$\gamma$ correlation analysis (where they would artifactually inflate the correlation by clustering at the origin) but retain them in the full condition table for completeness.

\section{Results}

\subsection{Condition Survey}

Table~\ref{tab:all} presents all 11 conditions. Five provide the complete $(\gamma, H, \text{CV})$ triple; two additional conditions provide $\gamma$ and CV (but lack weight vectors for $H$); four GPT-4o conditions provide $\gamma$ and $H$ (but the $H$ values are artifactual).

\begin{table}[t]
\centering
\caption{Complete condition survey. $\dagger$ GPT-4o conditions excluded from primary analysis (see \S\ref{sec:caveat}). $\ddagger$ Weight vectors not available for entropy computation.}
\label{tab:all}
\footnotesize
\begin{tabular}{@{}lccccc@{}}
\toprule
\textbf{Condition} & $\boldsymbol{\gamma}$ & $\boldsymbol{H}$ & $\textbf{CV}\boldsymbol{(N{=}5)}$ & $\boldsymbol{N}$ \\
\midrule
DS self-eval          & 0.033 & 0.992 & 2.420 & 30 \\
DS $\times$ Qwen      & 0.187 & 0.976 & 1.025 & 30 \\
DS self-eval r30$^\ddagger$ & 0.936 & ---   & 0.083 & 10 \\
Ablation max          & 1.038 & 0.753 & 0.157 & 10 \\
Qwen 3.7              & 1.059 & 0.793 & 0.108 & 8  \\
Ablation no-S0        & 0.979 & 0.788 & 0.161 & 5  \\
\midrule
GPT-4o replication$^\dagger$   & 0.000 & 1.000 & ---   & 8  \\
GPT-4o symmetric$^\dagger$     & 0.000 & 1.000 & ---   & 8  \\
GPT-4o checkpoint$^\dagger$    & 0.000 & 1.000 & ---   & 8  \\
GPT-4o sym check$^\dagger$     & 0.000 & 1.000 & ---   & 8  \\
GPT-4o DMXAPI$^\dagger$        & 0.000 & ---   & ---   & 10 \\
\bottomrule
\end{tabular}
\end{table}

\subsection{The Empirical Frontier}

Figure~\ref{fig:frontier} maps the five conditions with complete $(\gamma, \text{CV})$ metrics. Despite the limited sample, a clear structure emerges:

\begin{figure}[t]
\centering
\includegraphics[width=0.85\textwidth]{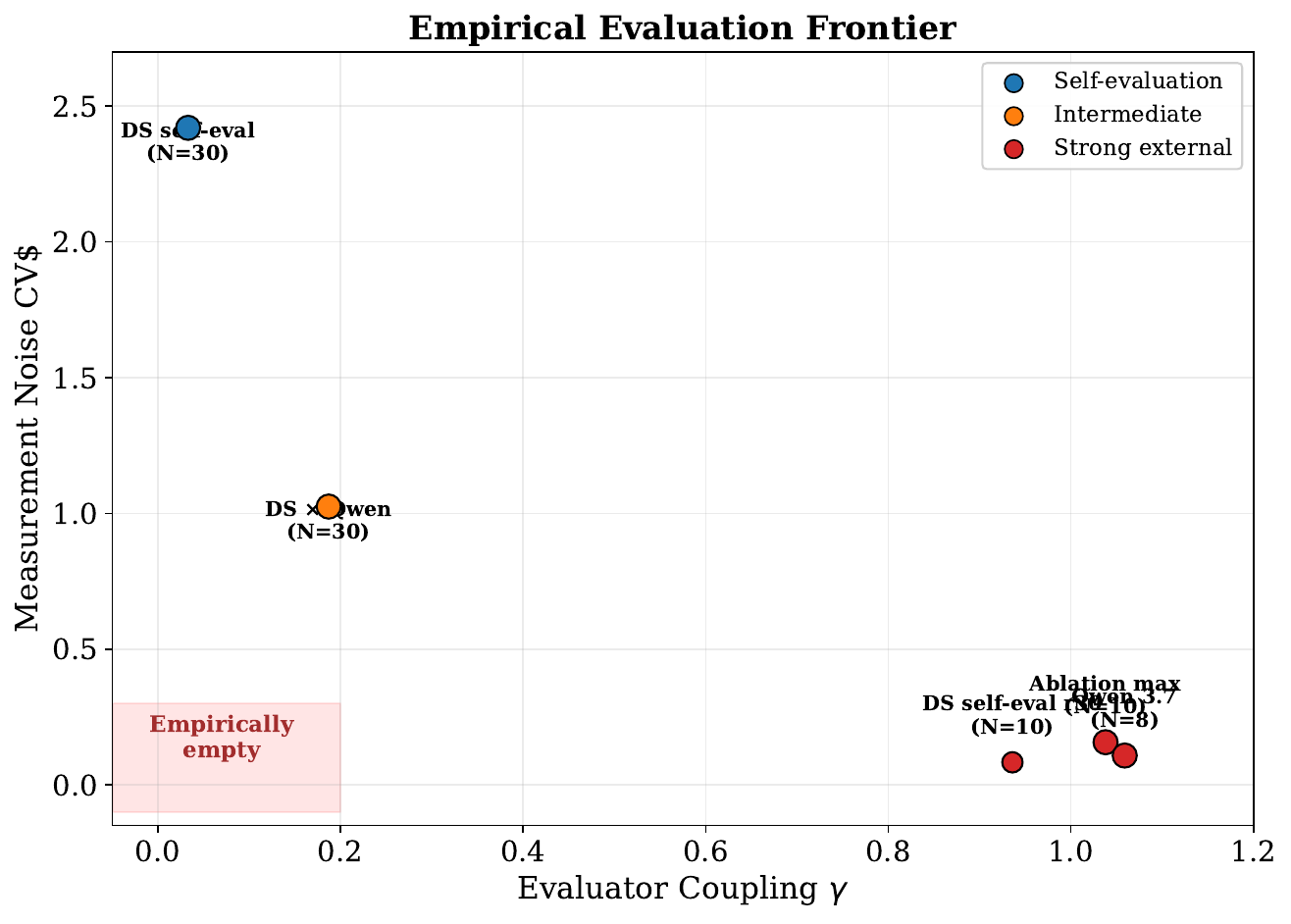}
\caption{The empirical evaluation frontier. Points show five conditions with complete $(\gamma, \text{CV})$ metrics. Color indicates strategy entropy $H$. The red-shaded region (low $\gamma$, low CV) is empirically empty.}
\label{fig:frontier}
\end{figure}

\textbf{Low-coupling regime} ($\gamma < 0.2$). DS self-eval ($\gamma = 0.033$, CV = 2.42) occupies the ``unbiased, unreliable'' corner. With near-zero evaluator coupling, measurement noise is extreme: the standard deviation of $\gamma$ estimates at $N{=}5$ is more than twice the mean.

\textbf{High-coupling regime} ($\gamma > 0.9$). DS self-eval r30, Ablation max, and Qwen 3.7 cluster at high $\gamma$ (0.94--1.06) and low CV (0.08--0.16). These conditions produce stable rankings---CV$(N{=}5) < 0.16$ in all cases---but the rankings primarily reflect evaluator preferences.

\textbf{Intermediate regime.} Only DS$\times$Qwen ($\gamma = 0.187$, CV = 1.025) occupies the transition zone between the two clusters. This regime is severely undersampled.

\textbf{Empty region.} The region $\{\gamma < 0.2, \text{CV}(N{=}5) < 0.3\}$ is empty. No evaluator--agent pair in our sample achieves both low bias and high reliability at $N{=}5$.

\subsection{Strategy Entropy Gradient}

Among the five conditions with valid $H$ measurements (excluding GPT-4o artifacts), entropy decreases with coupling: $r(H, \gamma) = -0.989$ ($p = 0.001$, $n = 5$). The DS self-eval condition exhibits near-maximal entropy ($H = 0.992$) under minimal coupling, while Ablation max shows substantially reduced entropy ($H = 0.753$) under strong coupling ($\gamma = 1.038$). The Ablation no-S0 condition ($\gamma = 0.979$, $H = 0.788$), with only $N{=}5$ seeds, provides an additional data point consistent with the trend.

\section{Discussion}

\textbf{The missing middle.} The empirical frontier is bimodal: conditions cluster at either very low or very high $\gamma$, with a sparsely sampled intermediate regime. This reflects current experimental practice---self-evaluation ($\gamma \approx 0$) and strong external evaluation ($\gamma > 0.9$) are the dominant paradigms. Deliberately designing evaluators with intermediate coupling (e.g., weak evaluators, ensemble evaluators with partial bias cancellation) would populate this regime and enable more precise characterization of the trade-off curve.

\textbf{GPT-4o version drift.} The four GPT-4o conditions all exhibit $\gamma = 0.000$ and $H = 1.000$---the evaluator exerts zero measurable influence on the agent's strategy distribution. This is consistent with the version drift documented in~\cite{liu2026mmepc}: GPT-4o's May 2026 version showed strong coupling ($\gamma \approx 1.176$), while the June 2026 version shows none. From the perspective of the trade-off, this positions GPT-4o as simultaneously the most ``unbiased'' and the least ``reliable'' evaluator---its rankings are uncorrelated with agent strategy, providing no signal for evaluation.

\textbf{Limitations.} Our survey has three principal limitations. First, all conditions come from a single research group's experiments, limiting generality. Independent replication with different models, tasks, and protocols is needed. Second, the sample size of five conditions with complete metrics is insufficient for reliable estimation of the trade-off curve's functional form. Third, the GPT-4o conditions produce degenerate metrics ($\gamma = 0$, $H = 1$) that may reflect API version artifacts rather than genuine evaluator behavior; these conditions should be re-measured with a stable API version or alternative evaluator models.

\textbf{Benchmark release.} We release all per-condition metrics as a standardized JSON dataset (\texttt{p16\_data.json} in supplementary material). Each entry contains the condition name, $\gamma$ mean and standard deviation, $H$ mean, standard deviation, and range, CV$(N{=}5)$, and number of seeds. We encourage the community to contribute additional evaluator--agent conditions to this benchmark using the standardized pipeline, following the model of multi-metric LLM evaluation established by~\cite{liang2023helm}.

\section{Conclusion}

An 11-condition empirical survey of the bias-reliability tradeoff confirms that evaluator coupling ($\gamma$) and measurement reliability (CV) are inversely related across diverse evaluator--agent pairs, with $r(H, \gamma) = -0.989$ ($n = 5$ complete conditions). The data reveal a bimodal empirical frontier---self-evaluation at the low-$\gamma$, high-CV extreme, strong external evaluation at the high-$\gamma$, low-CV extreme---with a sparsely sampled intermediate regime. GPT-4o's June 2026 version exhibits zero measurable evaluator coupling, consistent with documented version drift. All data are released as a public benchmark.

\section*{Broader Impact Statement}

This paper characterizes evaluator behavior using quantitative metrics. The framework could be misused to justify biased evaluation (``high $\gamma$ is acceptable because it improves reliability''), which we explicitly caution against: the trade-off should motivate larger sample sizes for unbiased evaluators, not acceptance of bias. The benchmark dataset may be used to compare evaluator models; such comparisons should account for API version effects (as demonstrated by the GPT-4o drift) and not be treated as stable over time.

\section*{Reproducibility Statement}

All data are drawn from the publicly available dataset of~\cite{liu2026mmepc}. The analysis pipeline (\texttt{triangle\_verification.py}) from~\cite{liu2026triangle} is included in the supplementary material. The per-condition benchmark dataset (\texttt{p16\_data.json}) is provided in machine-readable JSON format.

\end{document}